\documentclass[twoside]{article}

\usepackage[accepted]{style}

\usepackage[round]{natbib}

\usepackage{graphicx}
\usepackage{times}  
\usepackage{helvet}  
\usepackage{courier}  
\usepackage[hyphens]{url}  
\usepackage{graphicx} 
\usepackage[font=small]{caption}
\usepackage{amsmath} 
\usepackage{amssymb}
\usepackage{booktabs}
\usepackage{multirow}
\usepackage{rotating}
\usepackage{bm}
\usepackage{bbm}
\usepackage{hyperref}
\usepackage{multirow}
\usepackage{multicol}
\usepackage{amssymb}
\usepackage[capitalise]{cleveref}
\usepackage{array,multirow,graphicx}
 \usepackage{float}
\DeclareCaptionStyle{ruled}{labelfont=normalfont,labelsep=colon,strut=off} 
\usepackage{algorithm}
\usepackage{algorithmic}

\newcommand{\SR}{\operatorname{SR}}
\newcommand{\CIF}{\operatorname{CIF}}

\begin{document}

%
\runningtitle{On Training Survival Models with Scoring Rules}

%
\runningauthor{Kopper et al.}

\twocolumn[

\aistatstitle{On Training Survival Models with Scoring Rules}

\aistatsauthor{Philipp Kopper \And David Rügamer \And  Raphael Sonabend \And Bernd Bischl \And Andreas Bender }

\aistatsaddress{ \\ Department of Statistics, LMU Munich, \\ Munich, Germany \\
Munich Center for Machine Learning, \\ Munich, Germany  \And  \\ Imperial College London, \\ London, United Kingdom \And \\ Department of Statistics, LMU Munich, \\ Munich, Germany \\ 
Munich Center for Machine Learning, \\ Munich, Germany } ]

\begin{abstract}

Scoring rules are an established way of comparing predictive performances across model classes.
In the context of survival analysis, they require adaptation in order to accommodate censoring.
This work investigates using scoring rules for model training rather than evaluation.
Doing so, we establish a general framework for training survival models that is model agnostic and can learn event time distributions parametrically or non-parametrically.
In addition, our framework is not restricted to any specific scoring rule. 
While we focus on neural network-based implementations, we also provide proof-of-concept implementations using gradient boosting, generalized additive models, and trees.
Empirical comparisons on synthetic and real-world data indicate that scoring rules can be successfully incorporated into model training and yield competitive predictive performance with established time-to-event models.

\end{abstract}

\section{INTRODUCTION}
Survival analysis (SA) is an important branch of statistics and machine learning that deals with time-to-event data analysis. Let $Y>0$ be a random variable representing a time-to-event of interest (e.g., time-to-death after operation) and $y$ its realization. In many studies, $Y$ cannot be observed in all cases due to censoring $C>0$. Thus, in the presence of right-censoring, we can only observe realizations of $T:=\min(Y, C)$ and status indicator $D:=I(Y\leq C)$. Observed data is then given by tuples $(t_i, d_i, \mathbf{x}_i),i=1,\ldots,n$, where $t_i$ is an observed event or censoring time, $d_i$ the status indicator and $\mathbf{x}_i^\top = (x_{i1},\ldots,x_{ip})$ a $p$-dimensional feature vector.

Notably, while we are interested in inference about $Y$, we only have realizations of $T$. Therefore usual metrics for the evaluation of predictive performance based on the difference of true and observed value ($y_i-\hat{y_i}$) cannot be calculated for censored time-to-event data. For the same reason, most survival models don't generate predictions $\hat{y}$ but rather probabilistic predictions $\hat{F}_Y(\tau) = \mathbb{P}(Y\leq \tau), \tau \in \mathbb{R}_0^+$, or equivalently the survival function $\hat{S}_Y(\tau)=1-\hat{F}_Y(\tau)$. At the estimation stage, censoring needs to be accounted for in order to obtain unbiased estimates of ${S}_Y(\tau)$. Common approaches include parametric models that assume a specific distribution for the event times with a censoring-adjusted likelihood (e.g., accelerated failure time models) as well as non- and semi-parametric approaches that partition the follow-up into intervals and estimate the (baseline) hazard rate within each interval (e.g., Kaplan-Meier, Cox, discrete-time approaches).

For predictive modeling, dedicated evaluation metrics, that take into account the survival nature of the data, have been proposed in the literature (see \cite{sonabend2021} for an overview). Such metrics are often model-agnostic in order to allow comparison of predictive performances across model classes.
While concordance-based metrics \cite[e.g.\ Harrell's C,][]{harrell1982evaluating} are popular in practice, they only allow to evaluate of how well the model ranks the risk for an event. 
On the other hand, (strictly proper) scoring rules have been proposed as suitable tools to evaluate probabilistic (distribution) predictions \citep{gneiting2007strictly}.
As these scores often only rely on point-wise survival probability predictions (without for example requiring a density estimate), scores can be compared across different model classes.
One such scoring rule is the continuous rank probability score or integrated brier score \citep{gneiting2007strictly}. \cite{graf1999assessment} adapted it to the survival setting by weighting the scores with respect to the individuals' probabilities of being censored (IPCW). In this work, we refer to it as integrated survival brier score (ISBS). While ubiquitous in practice, recent work suggests that the ISBS is not proper \citep{rindt2022, sonabend2022proper, yanagisawaProperScoringRules2023}, and proper alternatives have been proposed. As scoring rules in survival analysis are established in the context of model evaluation and comparison, so far only few attempts have been made to use them as a loss function for model training.

\paragraph{Our Contributions}
In this work, we investigate the use of censoring-adapted scoring rules for model training rather than evaluation.
The developed framework uses gradient-based optimization of the scoring rule of choice, evaluated at discrete partitions of the follow-up. Two alternatives are proposed: estimation of parameters of an assumed distribution (\emph{parametric learning)} and direct estimation of the distribution (\emph{distribution-free approach}). The former generates a continuous time parametric distribution while the latter can be viewed as a variant of discrete time-to-event analysis applicable in the discrete-time setting but also in the continuous time setting via interpolation.
The approach is agnostic w.r.t.\ the choice of distribution (if parametric) as well as scoring rule. 
Additionally, while our main implementation is based on neural networks, we also show the applicability of our framework to other model classes like gradient boosting, trees, and generalized additive modeling.
We perform empirical evaluations of the approach on synthetic and real-world data, showing competitive predictive performance compared to established state-of-the-art survival models.

\section{RELATED LITERATURE} \label{sec:rellit}

\paragraph{Scoring rules} Scoring rules are established tools for model evaluation and comparison, particularly in the context of probabilistic predictions. A comprehensive summary is given in \cite{gneiting2007strictly} who also investigate the role of scoring rules for estimation. Adaptations of scoring rules for survival analysis (see Table \ref{tab:scoring-rules} for an overview of selected scores) have been pioneered by \cite{graf1999assessment}, who defined the integrated survival brier score (ISBS), which weights the integrated brier score by an estimate of the censoring distribution $\hat{G}$, usually using the Kaplan-Meier estimator. Other adaptations are discussed in \cite{dawid2014}, \cite{rindt2022}, \cite{yanagisawaProperScoringRules2023} and \cite{sonabend2022proper}. \cite{rindt2022} propose the right-censored log-likelihood (RCLL) and prove its properness, but its calculation requires an estimate of the density $f_Y$, which is not readily available for non- and semi-parametric methods that often only return survival probability predictions. The score proposed by \cite{yanagisawaProperScoringRules2023} is also proper but relies on an oracle parameter that is not known in practice. \cite{sonabend2022proper} suggest a class of re-weighted scoring rules (\cref{eq:reweighted-loss}), including the re-weighted ISBS (RISBS) and re-weighted integrated survival log-loss (RISLL):

\begin{equation}
\label{eq:reweighted-loss}
\SR_{R, i}(\tau) = \frac{d_i}{\hat{G}(t_i)}\SR_i(\tau),
\end{equation}
with $d_i$ being the status indicator, $\SR_i(\tau)$ is a suitable point-wise scoring rule (e.g.\ the IBS) of observation $i$ at time $\tau$ and $\hat{G}(t_i)$ an estimate of the censoring distribution at time $t_i$, which is estimated beforehand.
$\SR$ is computed up to a $\tau^* < \tau^{max}$, the largest observed survival time,  and
\cite{sonabend2022proper} recommend to consider all fully observed $i$ still at risk at $\tau^{*}$ with $d_i = 1$.

\paragraph{Survival Models} Most of the existing methods model the hazard function non- or semi-parametrically (i.e.\ without (strong) distributional assumptions) based on prior partitioning or discretization of the follow-up (for example Cox regression \citep{cox.regression.1972}, (extensions of) piece-wise exponential models \citep{Friedman1982} and discrete-time approaches \citep{tutzModelingDiscreteTimeEvent2016}), or use specific distributional assumptions with dedicated loss functions for censored data \citep[e.g.,][]{wei1992}. 
More recently, adaptations of these approaches based on machine and deep learning have been suggested \citep[cf.~][for respective reviews]{wang.machine.2019, wiegrebe2024deep}.
According to the latter, most deep learning models are adaptations of the Cox model, followed by discrete-time approaches. 
The latter are popular as they allow to transform a survival task to a classification task and don't require strong distributional assumptions. Concretely, the follow-up is partitioned into $J$ intervals $(\tau_{j-1}, \tau_j), j=1,\ldots,J;\ \tau_0 := 0$ and new status indicators are defined for each interval $d_{ij} = I(t_i \in (\tau_{j-1}, \tau_j] \wedge d_i = 1)$. Assuming a Bernoulli distribution for these new event indicators, discrete time methods than optimize the resulting Binomial log-likelihood. Popular methods within this class include DeepHit \citep{lee.deephit.2018} and \emph{nnet-survial} \citep{gensheimer.scalable.2019}.

In the context of SA, only few papers so far have suggested the use of scoring rules at the estimation or training rather than the evaluation step. A notable exception is \cite{avati2018}, who use a survival adapted continuous rank probability score (SCRPS), and \cite{rindt2022} who use RCLL for training. Both illustrate their approach using neural network-based implementations.
While \cite{avati2018} evaluate the SCRPS in order to estimate the parameters of a log-normal distribution, the approach by \cite{rindt2022} is distribution-free, but requires an estimate of the density $f_Y$ which is approximated in their implementation. The two approaches suggested in this work differ from previous endeavors. In contrast to other methods, our non-parametric approach learns increments of the survival function based on a scoring rule, whereas others use a specific likelihood. Our parametric approach is similar to \cite{avati2018} but not restricted to SCRPS or the log-normal distribution.

\section{TRAINING WITH SCORING RULES}
\label{sec:methods}

We aim to learn $F_Y$ through a discrete evaluation of an associated scoring rule.
While our approach is scoring rule agnostic, we focus on the rules in \cref{tab:scoring-rules}. The ISBS is of great historical significance and is a popular evaluation metric in the majority of benchmark experiments for SA. The alternatives in \cref{tab:scoring-rules} have been suggested only recently and therefore have not been applied often in practice. 
The SCRPS as implemented in \cite{avati2018} is the ISBS but without weighting contributions by inverse probability of censoring weights. The weighting factor in RISBS and RISLL means that contributions of censored observations are always set to zero and non-censored observations are censored by the probability of not being censored until the observed event time. 

\begin{table}[ht]
    \centering
    \scriptsize  
    \setlength{\tabcolsep}{2pt}  
        \caption{Selected model-agnostic scoring rules. Here, $F_i(\tau):=F(\tau|\mathbf{x}_i)$; $S_i$, $f_i$ equivalently. RCLL is only evaluated at the observed time $t_i$, while all other rules are evaluated over $[0, \tau^*]$. }
    \label{tab:scoring-rules}
    \renewcommand{\arraystretch}{1.1}  
    \begin{tabular}{>{\centering\arraybackslash}m{1.75cm} >{\centering\arraybackslash}m{6.25cm}}  
        \toprule
       \small \textbf{Abbreviation} \newline \tiny{(Source)}&  \small \textbf{Definition}\newline \ \\ \midrule
        \textbf{ISBS}\newline \citep{graf1999assessment} & 
        \small
        \(
        \begin{aligned}
         \int_{0}^{t_i} \frac{F_i(\tau)^2}{\hat{G}(\tau)}
          d\tau + \int_{t_i}^{\tau^*} \frac{d_i S_i(\tau)^2}{\hat{G}(t_i)} d\tau
        \end{aligned}
        \)
         \\ \midrule
        \textbf{SCRPS} \citep{avati2018} & 
        \small
        \(
        \begin{aligned}
          \int_{0}^{t_i} F_i(\tau)^2 d\tau  +
         \int_{t_i}^{\tau^*} S_i(\tau)^2 d\tau
        \end{aligned}
        \)
        \\ \midrule
        \textbf{RISBS} \citep{sonabend2022proper} & 
        \small
        \( 
        \begin{aligned}
        \int_{0}^{t_i} \frac{d_i F_i(\tau))^2}{\hat{G}(t_i)} d\tau + \int_{t_i}^{\tau^*} \frac{d_i S_i(\tau)^2}{\hat{G}(t_i)}
            \end{aligned}
            \)
         \\ \midrule
        \textbf{RISLL} \citep{sonabend2022proper} & 
        \small
        \(
        \begin{aligned}
       - \frac{d_i}{\hat{G}(t_i)} \Bigg(\int_{0}^{t_i} \log(F_i(\tau)) + \int_{t_i}^ {\tau^*}\log(S_i(\tau)) d\tau\Bigg)
        \end{aligned}
        \)
        \\ \midrule
        \textbf{RCLL} \citep{rindt2022} & 
        \small
        \(
        \begin{aligned}
          - \log  ( d_i f_i(t_i) + (1 - d_i) S_i(t_i)  )
        \end{aligned}
        \)
        \\ \bottomrule
    \end{tabular}
\end{table}

In order to use scoring rules for training, we partition the follow-up into $J$ equidistant intervals $(\tau_{j-1}, \tau_j], j = 1,\ldots,J$, with $\tau_0=0$ and $\tau_J$ the largest observed event time.
We then minimize the objective $O$ that evaluates scoring rule $\SR_i(\tau_j|\hat{G})$ for observation $i$ at time $\tau_j$, given censoring distribution $\hat{G}$:
\begin{equation}
\label{eq:total-loss}
O = \frac{1}{N}\sum_{i=1}^N \frac{1}{J}\sum_{j=1}^J \SR_{i}(\tau_j|\hat{G}).
\end{equation}
To do so, we need $J$ point-wise estimates of $S(\tau_j|\mathbf{x}_i) = 1 - F(\tau_j|\mathbf{x}_i)$.
These can be generated in two ways: 
\begin{itemize}
    \item[1.)]\emph{Parametric Learning}: Estimation of the parameters of an assumed distribution;
    \item[2.)] \emph{Distribution-free Approach}: Direct estimation of the survival function without distributional assumption.
\end{itemize}

\subsection{Modeling Approaches}

Both, approaches 1.) and 2.) share the same objective function \eqref{eq:total-loss} and only differ in the way the predictions $\hat{S}(\tau_j|\mathbf{x}_j)$ are obtained.
Both variants ensure that $\hat{S}$ is monotonically decreasing.
Details are given below. 

\paragraph{Parametric Learning}

One way to obtain estimates for $S(\tau|\mathbf{x}_j)$ is by assuming a parametric distribution of event times and learning the distribution's parameters.
Let $F(\tau|\boldsymbol{\theta})$ be a distribution suitable to represent event times $Y>0$, with parameters $\boldsymbol{\theta} \in \mathbb{R}^m$ depending on the input features, i.e., $\boldsymbol{\theta}(\mathbf{x}) = (\theta_1(\mathbf{x}), \theta_2(\mathbf{x}), \ldots, \theta_m(\mathbf{x}))^\top$. 
Some popular parametric survival distributions include the Weibull, log-logistic, and log-normal distribution. 
Given parameter estimates $\hat{\boldsymbol{\theta}}(\mathbf{x})$, all quantities of the distribution, including the survival function $\hat{S}(\tau|\hat{\boldsymbol{\theta}}(\mathbf{x})) = 1 - \hat{F}(\tau|\hat{\boldsymbol{\theta}}(\mathbf{x}))$, are fully specified and thus prediction can be obtained at any time point $\tau$.
Depending on the distribution, parameters may have restrictions, e.g.\ for the log-normal distribution $\boldsymbol{\theta}(\mathbf{x}) = (\mu(\mathbf{x}), \sigma(\mathbf{x}))^\top$ with $\mu \in \mathbb{R}$ and $\sigma \in \mathbb{R}_+$.
The distribution parameters $\boldsymbol{\theta}$ are learned by minimizing \cref{eq:total-loss} w.r.t.\ the model parameters. 

\paragraph{Distribution-free Approach}

Instead of obtaining an estimate of the survival function by learning the parameters of an assumed distribution, we can also learn the survival function by estimating the increments $\alpha_{i,j} := \alpha_{i,j}(\mathbf{x}_i) 
$ between the survival functions at subsequent discrete time points/intervals $\tau_{j-1}, \tau_j$. We require the following properties to obtain a valid survival function:
\begin{itemize}
\item[(a)] $S(\tau_{j}|\mathbf{x}_i)$ needs to be monotonically decreasing, i.e.\ $\alpha_{i,j}\leq 0$; 
\item[(b)] $S(\tau_{j}|\mathbf{x}_i) \in [0, 1]$;
\item[(c)] $\alpha_{i,j} \in [-1,0]$. 
\end{itemize}
In order to learn the increments $\alpha_{i,j}$, we require appropriate activation functions $\gamma_u(x) \in [0, 1], u \in \{1,2\}$, such as the sigmoid or truncated ReLU function $f(\cdot) = \min(1,\max(0,\cdot))$, and a model $g_l$ (e.g., a neural network) for the $l$th interval. By defining 
$$\hat{S}(\tau_{j}|\mathbf{x}_i)= 
    \gamma_2\left(\sum_{l=1}^j(-\gamma_1(g_l(\mathbf{x}_i))\right)
$$ 
through increments $\hat\alpha_{i,l} := -\gamma_1(\hat{g}_l(\mathbf{x}_i)) \in [-1,0]$, we obtain a monotonically decreasing survival function $\hat{S}(\tau_{j}|\mathbf{x}_i) = \gamma_2(\sum_{l=1}^j \hat\alpha_{i,l}) \in [0,1]$ for each time interval $\tau_j$ with $\tau_0 = 0$ and $\hat{S}(\tau_0|\mathbf{x}_i) = 1$.
In contrast to the parametric learning approach, this approach initially only produces discrete survival probabilities $\hat{S}(\tau_j|\mathbf{x})$. However, simple interpolation or smoothing \citep{archettiBridgingGapImprove2024} can be applied to obtain meaningful predictions at time points between initial interval points $\tau_j$.

\subsection{Competing Risks}

In the competing risks setting, we are interested in the time until the first of $K$ competing events is observed. Let $E \in \{1,\ldots,K\}$ be a random variable representing the possible event types with realizations $e$. In this setting, we are typically interested in estimating $P(Y\leq \tau, E=e|\mathbf{x})$, i.e., the probability of observing an event of type $e$ before time $\tau$ given feature set $\mathbf{x}$. This quantity is usually referred to as cumulative incidence function (CIF) and denoted by $\CIF_k(\tau|\mathbf{x}), k=1,\ldots, K$. 

In the case of our parametric framework, we either learn the set of parameters for each competing risk $k$ with separate sub-models for distribution parameters $\bm{\theta}_k$ or train a single joint model for all parameters $\mathbf{\Theta} = \{\bm{\theta}\}_{k=1}^K$. 

The CIF in the non-parametric case is modeled via
\begin{equation} \label{eq:CIF}
    \widehat{\CIF}_k(\tau_{j}|\mathbf{x}_i) = \gamma_2\left(\sum_{l=1}^j(\gamma_1(g_{l,k}(\mathbf{x}_i))\right),
\end{equation}
where $g_{l,k}$ are now interval- and risk-specific models.

In order to evaluate competing risk models, we can use the single-risk scoring rules, but need to define a cause-specific status indicator 
\begin{equation}
\label{eq:cr_indicator}
d_{i,k} = d_i \mathbbm{1}(e_i=k) \in \{0, \ldots, K\},
\end{equation}
where $e_i$ is the cause observed for subject $i$.
We further constrain $\sum_{k=1}^K \hat{F}_k(\tau_{j}|\mathbf{x}_i) \leq 1$.
This can be achieved through the network architecture or by reweighting the resulting CIFs.
Putting everything together, we optimize the competing risks objective
\begin{equation}
\label{eq:total-loss-cr}
O^{\text{\tiny CR}}= \sum_{k=1}^K\frac{1}{N}\sum_{i=1}^N\frac{1}{J}\sum_{j=1}^J \SR_{i,k}(\tau_j|\hat{G}),
\end{equation}
where $\SR_{i,k}$ is a single-event scoring rule (e.g.\ \cref{tab:scoring-rules}) with the status indicator $d_i$ replaced by the competing risks indicator from \cref{eq:cr_indicator}.
Depending on the scoring rule, predictions can be directly obtained from the model or are internally reweighted.

\subsection{Optimization and Implementation}

\paragraph{Gradient-based Optimization}
For all scoring rules discussed in this paper, first derivatives with respect to an arbitrary weight vector $\omega$ of $S(\tau|\mathbf{x}_i, \omega)$ or $F(\tau|\mathbf{x}_i, \omega)$ exist. 
For example, for RISBS and a single observed observation $i$ and interval $j$
$$\frac{\partial}{\partial \omega} \SR(\tau; \hat{G}) \propto  F(\tau|\mathbf{x}_i, \omega) 
\frac{\partial}{\partial \omega} F(\tau|\mathbf{x}_i, \omega)$$
if $\hat{G}(t_i)$ is considered constant, which is usually the case if it is determined \emph{a priori}.
If $\hat{S}(\tau|\mathbf{x}_i)$ is differentiable itself, which is typically the case for neural networks, the model itself is differentiable. 

\begin{figure}
\centering
\includegraphics[width=0.75\linewidth]{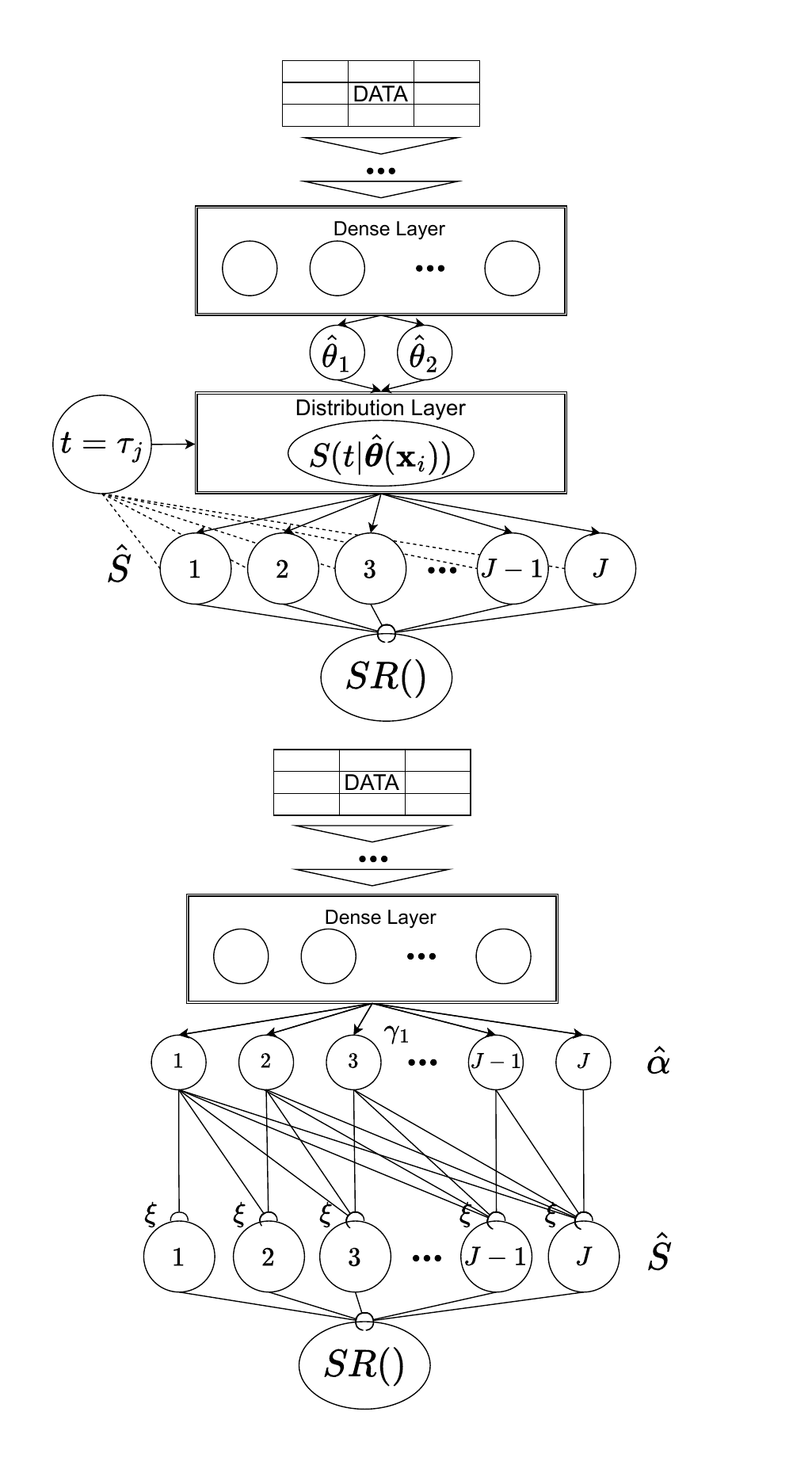}
\caption{\small Examples for architectures of our proposed method in the single risk case.
Top: Parametric approach. We pass the data through a fully connected neural network to estimate the parameters (here $\theta_1$ and $\theta_2$) of a survival distribution. We generate predictions for each $t = \tau_j$ using the parameterized $F$. Bottom: Non-parametric approach. We pass the data through a fully-connected neural network to estimate the survival increments $\alpha_j$ and use them to generate survival predictions for each $\tau_j$, where $\xi(\cdot) := \gamma_2(-\sum(\cdot))$.}
\label{fig:arch}
\end{figure}

\begin{figure}
    \centering
    \includegraphics[width=0.99\linewidth]{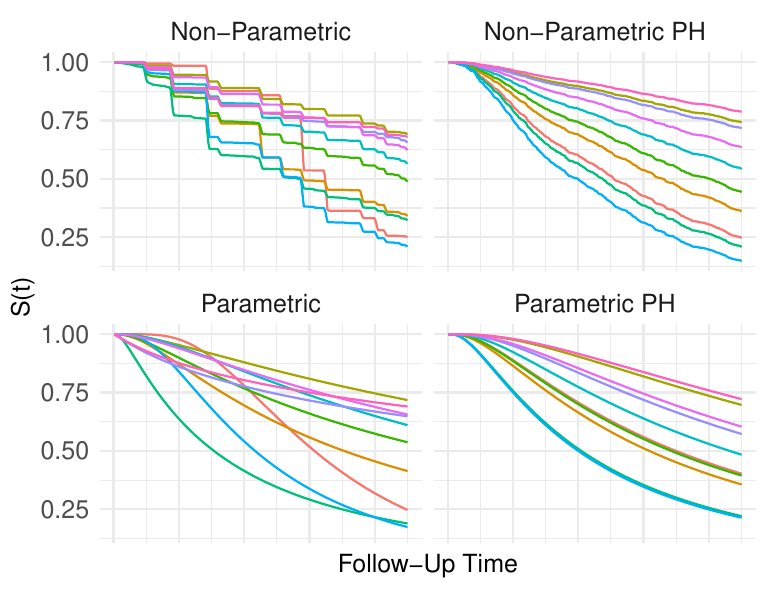}
    \caption{Selected predictions from the parametric and non-parametric framework for the \emph{metabric} data set. All models are tuned. 
    The variants relying on the PH assumption do not allow curves to cross. The step functions for the non-parametric framework have a varying roughness depending on the parametrization.}
    \label{fig:examples}
\end{figure}

\paragraph{Implementation}
Our framework can be easily implemented in a neural network. The network trunk can have an arbitrary shape whereas the output layer contains $J\times K$ units for both the parametric and non-parametric variant. The final layer of the parametric variant is a deterministic distributional layer that automatically enforces monotonicity by implementing the cumulative distribution or survival function of a parametric distribution. 
By the chain rule, backpropagation is given by the derivative of the $\SR$ w.r.t.~the parameters ${\boldsymbol{\theta}}_k$ of the chosen survival distribution for each risk $k$ times the gradients of ${\boldsymbol{\theta}}_k$ w.r.t.~the networks weights. 
For the non-parametric version, the output layer is specified as in \cref{eq:CIF}.
Exemplary architectures are shown in Figure \ref{fig:arch}.
Overfitting can, e.g., be addressed by dropout layers throughout the network architectures and L2 regularization on the ultimate layer's weights.
The parametric framework produces smooth continuous-estimates, the non-parametric one interpolated step-functions as can be observed in Figure \ref{fig:examples}.
In many cases, it is reasonable not to choose $\tau^* = \tau^{max}$ but slightly smaller (e.g.\ the 80th or 90th percentile) as late events have outlier character in small data sets \citep{sonabend2022proper}.

\paragraph{Alternative Implementations}
While the most versatile implementation is achieved through neural networks, the idea can be generalized to arbitrary machine learner models.
Particularly, the parametric framework with RISLL or ISLL as a loss function is applicable to some established machine learning models without further modification.
For an assumed lognormal or log-logistic distribution with location $\mu$ and scale $\sigma$, we only need to model the linear predictor $\log(\tau) / \sigma - \mu / \sigma$, apply a logit (log-logistic) or probit (lognormal) link, and optimize a weighted binary cross-entropy loss.
As $\sigma > 0$, monotonicity must be enforced on the estimation of $\frac{1}{\sigma}$, either implicitly (linear models) or explicitly (XGBoost).
Furthermore, we can use distributional regression software 
to generate predictions for any model class, e.g.\ trees, independent of their optimization.
Details on prototypical implementations of generalized additive models (GAM), XGBoost and
a distributional soft regression tree
are given in the Supplement.

\section{NUMERICAL EXPERIMENTS}

In the following sections, we evaluate our framework empirically. 
First, we test our approach with simulated data.
While our proposed method can represent arbitrarily complex associations, our goal is to show that the proposed method can estimate parametric and semi-parametric SA methods that traditionally optimize likelihoods:
Accelerated Failure time models (AFT) and the Cox proportional model hazards (CPH).
Furthermore, we explore how well our framework performs on benchmark data sets commonly used in SA for both single risk and competing risks.
For a meaningful comparison, we benchmark our neural network implementation against other deep learning algorithms.
However, we also include the oblique random survival forest \citep[ORSF;][]{jaeger.oblique.2019}, which has been shown to yield good predictive performance in SA tasks.
Last, we illustrate that the framework is also applicable to learners different from neural networks.

\paragraph{Evaluation and Tuning}
In all experiments, we make use of repeated subsampling.
For all benchmark data sets, we repeat the subsampling 25 times, and, except for KKBox, use 80\% of the data for training and 20\% for evaluation.
Repeated subsampling is preferred over cross-validation as the test set needs to be sufficiently large to estimate the censoring probability for all evaluation metrics. 
The number of subsamples depends on the complexity of the underlying experiments.
For KKBox, 2 percent of the data (ca.\ 1,000 events) is sufficient for model evaluation.
If tuning is necessary, we use a random search with a budget of 25 configurations. 
The inner loop of the nested resampling is a five-fold cross-validation.
Early stopping is performed when necessary based on the validation error.
In all experiments, models are evaluated using the RISBS as our main evaluation metric at different quantiles (25, 50, and 75 percent) of the follow-up.

\subsection{Comparison to Maximum Likelihood Estimation}
\label{ssec:sim-aft}
We first empirically check whether our approach can recover the parameters of a known event time distribution without explicitly using its likelihood for estimation. 
We compare the goodness of approximation with the true parameters and the ones estimated through maximum likelihood.
To do so, we simulate event times from an AFT model via
\begin{equation}
\label{eq:sim-aft}
\log(T_i) = \beta_0 + \beta_1 x_1 + \beta_2 x_2 + \beta_3 x_3 + \theta_2 \epsilon_i
\end{equation}
with $\boldsymbol{\beta}^\top = (2, 0.5, 0.2, 0)$ and let $\epsilon$ follow the (i) Logistic, (ii) Normal, and (iii) Extreme value distribution, implying event times $T\sim F(\theta_1(x_1,x_2,x_3), \theta_2)$ that follow a (i) Loglogistc, (ii) Log-normal and (iii) Weibull distribution, respectively. 
We only let one parameter of the distribution depend on features, i.e.\ $\theta_1(x_1, x_2, x_3) = \beta_0+ \beta_1 x_1 + \beta_2 x_2 + \beta_3 x_3$ and set $\theta_2 = \sigma = 0.4$.

\begin{figure*}
    \includegraphics[width=1\linewidth]{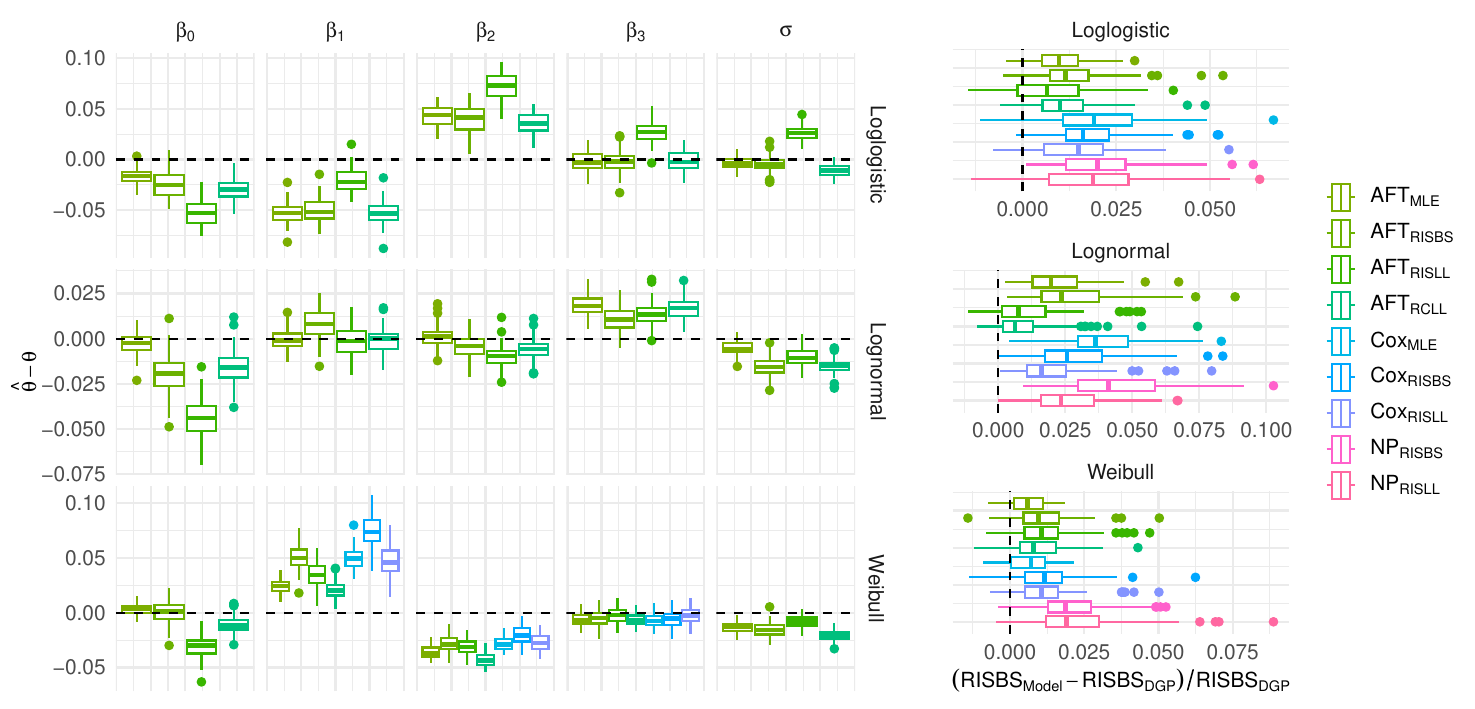}
    \caption{Results of the comparison to ML estimation. \textbf{Left}: difference of estimated parameter $\hat{\theta}$ to oracle parameters $\theta$. Parameter comparison for Cox PH models is limited to the coefficients $\beta_1, \beta_2$ and $\beta_3$, and the Weibull distribution. 
    \textbf{Right}: Relative difference in the predictive performance w.r.t.\ the data generating process (DGP). Optimal performance is given by $\text{RISBS}_{\text{DGP}}$, obtained by using true parameters in the correctly specified model. 
    }
    \label{fig:sim-aft}
\end{figure*}

For the simulation, we draw $n=1500$ event times from each distribution based on \cref{eq:sim-aft} and introduce censoring assuming a uniform distribution over the follow-up, resulting in approximately 28\% censoring.
We repeat this $B=100$ times, each time splitting the data into train (80\%) and test (20\%) data. 
In each iteration we calculate  
$$
\beta_j - \hat{\beta}_{j,m};\ j=0,\ldots,3; $$
$$\ m \in \{\text{AFT}_{\text{MLE}}, \text{Cox}_{\text{MLE}}, \text{AFT}_{\text{SR}}, \\ 
\text{Cox}_{\text{SR}}\},
$$
where we either optimize the respective correctly specified AFT models and Cox PH models via maximum likelihood estimation (MLE) or via one of the scoring rules (SR) based approaches as proposed in Section~\ref{sec:methods} . 
Additionally, we consider the difference between the estimated and true scale $\hat{\sigma} - \sigma$.
Next to the recovery of coefficients, we report the aggregated predictive performance of all models, in terms of the RISBS, RISLL, and ISBS.
For predictive performance evaluation, we also fit the non-parametric variant of our framework, $\text{NP}_{\text{SR}}$ with SR $\in\{$RISBS, RISLL$\}$.
The AFT model estimated within our framework with RCLL provides a direct comparison to $\text{AFT}_{\text{\tiny MLE}}$.
This results in a total of 4 AFT models (3 SR and 1 MLE), 2 Cox PH models (2 SR and 1 MLE), and 2 non-parametric models (both SR) for the main analysis.

The experimental results are presented in Figure \ref{fig:sim-aft}. 
The methods specified within our framework recover the true coefficients well with, however, a little approximation error.
This approximation error is negligibly small when considering the predictive performance in the right panel.
This finding holds for both, AFT and Cox PH model.
When considering the model performances, we also see that a simple un-tuned, yet regularized, non-parametric scoring-rule-based method performs comparably to the other (correctly specified) methods.
In contrast to other deep AFT (e.g.\ \citet{avati2018}) approaches, our method allows the estimation of a variety of distributions using the parametric framework, including the Weibull distribution which has repeatedly been reported to suffer from poor computational conditioning \citep[e.g.,][]{avati2018}.

\subsection{Benchmark Study}

In this section, we evaluate the models' predictive performance on synthetic and real-world data for both, single-event (\cref{tab:resultsBM}) and competing risks (\cref{tab:resultsBM2}) settings.

\begin{table*}
    \centering
\caption{Predictive performance of different learning algorithms for different data sets for single event using the RISBS (smaller is better). We report the mean and standard deviation (in brackets) from 25 distinct train-test splits and highlight the best method in \textbf{bold}. The AFT models are also tuned w.r.t.\ the distribution family. 
} \label{tab:resultsBM}
\centering
  \resizebox{0.98\textwidth}{!}{
\begin{tabular}{ccccccccccc}
\toprule
 &  & KM & Cox PH & ORSF & DeepSurv & nnet & DeepHit & $\text{AFT}_{\text{\tiny SCRPS}}^{\text{\tiny deep}}$ & $\text{AFT}_{\text{\tiny RISBS}}^{\text{\tiny deep}}$ &  $\text{NP}_{\text{\tiny RISBS}}$\\
\midrule
\underline{tumor} & Q25 & 7.4 (1.58) & 6.6 (1.43) & 6.5 (1.23) & 6.6 (1.52) & 6.5 (2.03) & 6.7 (1.80) & \textbf{6.4} (1.30) & 6.5 (1.37) & 6.6 (1.30)\\
$n=776$ & Q50 & 13.0 (1.46) & 11.7 (1.46) & 11.8 (1.40) & 11.6 (1.51) & 11.5 (1.84) & 11.6 (1.91) & 11.4 (1.21) & \textbf{11.3} (1.24) & \textbf{11.3} (1.32)\\
$p=7$ & Q75 & 17.8 (1.17) & 16.3 (1.47) & 16.4 (1.40) & 16.3 (1.40) & 16.2 (1.37) & 16.2 (1.67) & 16.2 (1.29) & \textbf{16.1} (1.30) & \textbf{16.1}  (1.39)\\
\hline \\[-1.8ex] 
\underline{gbsg2} & Q25 & 4.9 (0.80) & 4.7 (0.75) & \textbf{4.6} (0.73) & 4.7 (0.80) & 4.9 (1.03) & \textbf{4.6} (1.01) & 4.7 (0.75) & 4.7 (0.69) & \textbf{4.6} (0.80)\\
$n=2232$ & Q50 & 10.1 (0.80) & 9.3 (0.69) & 9.2 (0.73) & 9.3 (0.82) & 9.2 (0.95) & 9.5 (0.83) & 9.1 (0.69) & 9.4 (0.68) & \textbf{9.1} (0.79)\\
$p=7$ & Q75 & 15.4 (0.56) & 13.9 (0.47) & 13.6 (0.60) & 13.7 (0.62) & 13.6 (0.82) & 13.4 (0.71) & 13.4 (0.61) & 13.3 (0.54) & \textbf{13.2} (0.61)\\
\hline \\[-1.8ex] 
\underline{metabric} & Q25 & 5.1 (0.64) & 5.0 (0.61) & 5.1 (0.56) & 5.0 (0.60) & 5.5 (0.51) & 5.2 (0.55) & \textbf{4.9} (0.61) & \textbf{4.9} (0.56) & 5.0 (0.65)\\
$n=1904$ & Q50 & 11.4 (0.80) & 10.8 (0.78) & 10.9 (0.67) & 10.7 (0.60) & 10.9 (0.58) & 10.9 (0.70) & \textbf{10.4} (0.73) & \textbf{10.4} (0.72) &\textbf{10.4} (0.79)\\
$p=9$ & Q75 & 16.5 (0.61) & 15.2 (0.60) & 15.8 (0.66) & 15.1 (0.55) & 15.1 (0.46) & 15.4 (0.52) & 15.0 (0.61) & \textbf{14.8} (0.55) & \textbf{14.8} (0.59)\\
\hline \\[-1.8ex] 
\underline{breast} & Q25 & 2.2 (1.14) & 2.2 (1.14) & 2.1 (1.11) & 2.3 (1.13) & 2.2 (1.01) & 2.2 (1.02) & \textbf{2.0} (0.86) & 2.1 (0.97) & 2.4 (1.12)\\
$n=614$ & Q50 & 4.6 (1.51) & 4.6 (1.51) & \textbf{4.3} (1.44) & 4.8 (1.59) & 4.4 (1.35) & 4.5 (1.42) & 4.4 (1.18) & 4.4 (1.26) & 4.4 (1.54)\\
$p=1690$ & Q75 & 7.8 (1.62) & 7.8 (1.62) & \textbf{7.0} (1.61) & 7.6 (1.71) & 7.2 (1.60) & 7.3 (1.64) & 7.2 (1.50) & 7.3 (1.52) & 7.3 (1.75)\\
\hline \\[-1.8ex] 
\underline{KKBox} & Q25 & 1.02 (0.05) & 0.92 (0.04) &  & 0.87 (0.05) & 0.95 (0.07) & 0.93 (0.06) & \textbf{0.85} (0.06) & 0.86 (0.05) & 0.90 (0.07)\\
$n=865$ K & Q50 & 1.66 (0.06) & 1.41 (0.05) & -- & 1.25 (0.05) & 1.31 (0.06) & 1.35 (0.06) & 1.27 (0.05) & 1.21 (0.05) & \textbf{1.20} (0.07)\\
$p = 6$ & Q75 & 2.72 (0.09) & 2.19 (0.07) &  & \textbf{1.89} (0.06) & 2.00 (0.05) & 2.01 (0.06) & 1.94 (0.06) & 1.92 (0.06) & \textbf{1.89} (0.06)\\
\hline \\[-1.8ex] 
\underline{synthetic} & Q25 & 6.7 (0.70) & 4.7 (0.51) & 4.2 (0.66) & 3.5 (0.63) & 4.4 (0.66) & 3.9 (0.61) & \textbf{3.3} (0.48) & 3.4 (0.40) & 3.5 (0.56)\\
$n=1500$ & Q50 & 13.9 (0.78) & 9.2 (0.53) & 8.8 (0.81) & 6.7 (0.78) & 8.5 (0.71) & 8.0 (0.69) & \textbf{6.4} (0.36) & \textbf{6.4} (0.38) & \textbf{6.4} (0.51)\\
$p=4$ & Q75 & 19.7 (0.26) & 12.8 (0.56) & 10.0 (1.01) & 9.2 (0.76) & 9.6 (0.83) & 9.5 (0.67) & 8.6 (0.45) & \textbf{8.3} (0.46) & 8.4 (0.50)\\
\bottomrule
\end{tabular}
}
\end{table*}

\subsubsection{Single Risk}

We compare our framework to other popular deep learning models for survival analysis, namely nnet-survival \citep{gensheimer.scalable.2019}, DeepHit \citep{lee.deephit.2018}, and DeepSurv \citep{katzman2018deepsurv}, as well as the Countdown model (with an assumed log-normal distribution) as proposed in \citet{avati2018} ($\text{AFT}_{\text{\tiny SCRPS}}^{\text{\tiny deep}}$). 
For our framework we fit both, a non-parametric version $\text{NP}_{\text{\tiny RISBS}}$ and a deep parametric variant $\text{AFT}_{\text{\tiny RISBS}}^{\text{\tiny deep}}$.
Furthermore, we compare with baselines (KM and CPH) and Oblique Random Survival Forests (ORSF). All methods have been tuned over 50 configurations (10 for KKBox) except for the KM and CPH baselines. $\text{AFT}_{\text{\tiny SCRPS}}^{\text{\tiny deep}}$ approximates the model proposed in \cite{avati2018} (log-normal distribution, SCRPS as scoring rule). The model by \cite{avati2018} itself suffered from computational issues and did not result in adequate predictive performance. While not being perfectly identical to \cite{avati2018} $\text{AFT}_{\text{\tiny SCRPS}}^{\text{\tiny deep}}$ adapts their idea. 

We selected common data sets in the survival analysis literature primarily related to various medical conditions with observations in the high hundreds or low thousands and a large churn data set: tumor \citep{pammtools}, gbsgs2 \citep{schumacher1994randomized}, metabric \citep{curtis2012genomic}, breast \citep{ternes.identification.2017} and mgus2 \citep{kyle.long-term.2002}.
{KKBox} \citep{kkbox} is a large churn data set obtained from Kaggle that we processed for SA.
For {KKBox}, ORSF evaluations, however, failed due to the size of the data set. For similar computational reasons, Cox PH only uses one (non high-dimensional) feature for \emph{breast}.
\paragraph{Results} 
In summary, the results indicate that our proposed methods provides good predictive performance, competitive with established methods.
We observe that for the AFT model, both scoring rules (SCRPS and RISBS) have very similar performances.
This finding is in line with \cite{sonabend2022proper} who empirically study differences between proper and improper scoring rules and report only small differences.
$\text{NP}$ tends to perform worst on early quantiles indicating potential overfitting for the early cut points.
When using the ISBS as evaluation metric (given in the Supplement), similar results are obtained.

\subsubsection{Competing Risks}

\begin{table}[t]
\small
    \centering
\caption{\small Prediction accuracy of different methods (columns) for different competing risks data sets (rows) evaluated using the ISBS (smaller is better). We report the mean and standard deviation (in brackets) from 25 distinct train-test splits and highlight the best method in \textbf{bold}. 
}
  \label{tab:resultsBM2} 
  \resizebox{1.0\columnwidth}{!}{
\begin{tabular}{@{\extracolsep{1pt}} ccccccc} 
 & & AJ & CR PAMM & DeepHit  & $\text{AFT}_{\text{\tiny ISBS}}^{\text{\tiny deep}}$ \\ 
\hline \\[-1.8ex] 
\underline{mgus2}& Q25 & \textbf{1.1} (0.39) & \textbf{1.1}  (0.39) & \textbf{1.1} (0.42)  & 1.2 (0.58) \\
(cause 1) & Q50 & 2.1 (0.59) & \textbf{2.0} (0.57) & 2.2 (0.59)  & 2.2 (0.58) \\
$n=1384$ & Q75 & 3.2 (0.84) & 3.2 (0.80) & \textbf{3.1} (0.85)  & 3.3 (0.79) \\
\hline \\[-1.8ex] 
\underline{mgus2} & Q25 & 9.1 (1.18) & \textbf{8.6} (1.16) &  8.7 (1.04) & 8.7 (1.10) \\
(cause 2) & Q50 & 14.3 (1.03) &  13.2 (1.17)  & \textbf{12.9} (1.20)  & 13.0 (1.34) \\
$p = 6$ & Q75 & 18.2 (0.74) & 15.8 (0.99) & \textbf{15.6} (1.07)  & 15.7 (1.22) \\
\hline \\[-1.8ex] 
\underline{synthetic} & Q25 & 5.4 (0.66) & 3.7 (0.49) & 3.2 (0.56)  & \textbf{3.1} (0.45) \\
(cause 1) & Q50 & 10.9 (0.96) & 7.3 (0.67) & \textbf{5.7} (0.69)  & \textbf{5.7} (0.60) \\
$n=1500$ & Q75 & 16.9 (0.83) & 11.5 (0.55) & \textbf{8.3} (0.77)  & 8.4 (0.65) \\
\hline \\[-1.8ex] 
  \underline{synthetic}  & Q25  &  2.2 (0.72)   & \textbf{2.0}  (0.59)   & \textbf{2.0} (0.62) & 2.1 (0.53) \\
(cause 2)& Q50 & 5.8 (0.97) & \textbf{4.7} (0.78) & 5.0 (0.79)  & 4.9 (0.73) \\
$p = 4$ & Q75 & 9.7 (1.06) &  7.9 (0.94) & 7.9 (1.00)  & \textbf{7.8} (0.79) \\
\hline \\[-1.8ex] 

\end{tabular} 
}
\end{table}

For competing risk, we compare our approach against the baseline methods Aalen-Johannsen estimator \citep[AJ;][]{aalen.nonparametric.1978} and competing risks piecewise exponential additive model \citep[CR PAMM;][]{hartl2022protein} as well as DeepHit, which is typically considered when dealing with competing risks. 
Next the the real-world data set \emph{mgus2}, we also consider a synthetic data set with a risk with complex (cause 1) and simple (cause 2) feature associations. 

\paragraph{Results} DeepHit and our method perform similarly well in estimating survival probabilities in a competing risk setting.
In some cases, both methods cannot outperform a CR PAMM that assumes linear effects on the log hazards.
However, this is most likely due to the differences between the empirical incidence of the two causes (only 12 \% of observed events in \emph{mgus2} are due to cause 1) and by construction (cause 2 of \emph{synthetic} assumes only linear associations).

\subsubsection{Alternative Implementations}

To test alternative implementations of our framework, we use two datasets: \emph{simple} is the simulation introduced in \cref{ssec:sim-aft} and reflects linear effects only, while \emph{complex} uses four features that partially exhibit non-linearities and interactions. 
Both settings assume a log-normal distribution.
As learners, we include KM and Cox PH for baseline comparisons and prototype implementations for an XGBoost, GAM, and soft regression tree. 
Performance is reported quantile-wise and estimated with 25 times repeated subsampling (80-20).
Methods are not tuned as our goal is only to show that these alternatives work in principle.
We find that all methods perform better than the baseline (KM). 
A single tree, however, does not manage to outperform a Cox PH model.
In the simple setting, GAM performs best. 
This is because the Cox PH model is slightly misspecified in the presence of log-normally distributed survival times.
Unsurprisingly, the (untuned) XGBoost model overfits in this simple regime.
For the complex setting, we allowed the GAM to capture non-linearities yet no interactions.
While this significantly boosts performance over Cox PH, the XGBoost approach manages to achieve very good generalization despite being unturned. All in all, this suggests that the methods work as intended and provide reasonable results.

\begin{table}[t]
\small
    \centering
\caption{Comparison of predictive performances of different learners (columns) for different datasets (rows). We report the mean (top) and standard deviation (below in brackets) from 25 distinct train-test splits. The best method is highlighted in \textbf{bold}. }
  \label{tab:xxxxx} 
   \resizebox{1.0\columnwidth}{!}{
\begin{tabular}{cccccccccc} 
&
&  KM 
&  Cox PH 
& $\text{AFT}_{\text{\tiny RISLL}}^{\text{\tiny GAM}}$ 
&  $\text{AFT}_{\text{\tiny RISBS}}^{\text{\tiny tree}}$
& $\text{AFT}_{\text{\tiny RISLL}}^{\text{\tiny XGB}}$   \\ 
\hline \\[-1.8ex] 
\underline{simple} & Q25 & 5.1 & 3.2 & \textbf{3.0} & 4.5 & 3.2 \\
                &      & (0.46) & (0.46) & (0.43) & (0.87) & (0.42) \\
$n=1500$& Q50 & 10.7 & 6.3 & \textbf{6.1} & 8.9 & 6.3 \\
                &      & (0.87) & (0.43) & (0.40) & (0.69) & (0.43) \\
$p=3$& Q75 & 15.5 & 8.7 & \textbf{8.5} & 13.3 & 8.9 \\
                &      & (0.42) & (0.51) & (0.44) & (0.60) & (0.51) \\
\hline \\[-1.8ex] 
\underline{complex} & Q25 & 6.7 & 4.7 & 3.8 & 6.2 & \textbf{3.6} \\
                &      & (0.71) & (0.52) & (0.54) & (0.59) & (0.47) \\
$n=1500$& Q50 & 13.9 & 9.2 & 7.2 & 12.7 & \textbf{6.6} \\
                &      & (0.78) & (0.53) & (0.59) & (0.62) & (0.44) \\
$p=4$& Q75 & 19.7 & 12.8 & 9.6 & 18.2 & \textbf{8.6} \\
                &      & (0.26) & (0.56) & (0.51) & (0.51) & (0.40) \\
\hline \\
\end{tabular} 
 }
\end{table}

\subsection{Ablation: Altering the Scoring Rule}

As discussed in \cref{sec:methods}, our framework is scoring-rule agnostic. 
While we mainly focused on RISBS in the experiments, we also implemented all other scoring rules from \cref{tab:scoring-rules}.
To investigate their influence, we study how the results from the benchmarking study qualitatively change when the optimized scoring rule is changed for the parametric sub-framework.

\begin{table}[h]
    \centering
\caption{Predictive performance of the ablation study. For two data sets from the benchmark study, we changed the training scoring rule to RISLL, RCLL and ISBS, respectively. 
For RISLL and ISBS we only consider the 75 \% percentile for $\tau^*$.
We also evaluate using the same scoring rules.
RCLL is reported as -RCLL.
} \label{tab:resultsBM3}
\centering
  \resizebox{1.0\columnwidth}{!}{
\begin{tabular}{cccccc}
 &  & $\text{AFT}_{\text{\tiny RISBS}}^{\text{\tiny deep}}$ & $\text{AFT}_{\text{\tiny RISLL}}^{\text{\tiny deep}}$ & $\text{AFT}_{\text{\tiny RCLL}}^{\text{\tiny deep}}$ & $\text{AFT}_{\text{\tiny ISBS}}^{\text{\tiny deep}}$ \\
\midrule
\underline{gbsg2} & RISBS & 13.6 (0.54) & \textbf{13.4} (0.59) & 13.5 (0.58) & 13.6 (0.56)\\
 & RISLL & 40.8 (1.39) & \textbf{40.7} (1.35) & 40.8 (1.47) & 41.2 (1.55)\\
$n=2232$ & RCLL & 2.67 (0.76) &  2.65 (0.72) & \textbf{2.60} (0.70) & 2.69 (0.79) \\
$p=7$ & ISBS & 13.1 (0.53) &  13.1 (0.59) & \textbf{13.0} (0.62) & 13.2 (0.63)\\
\hline \\[-1.8ex] 
\underline{synthetic} & RISBS & \textbf{8.3} (0.46) & \textbf{8.3} (0.50) & 8.4 (0.45) & 8.6 (0.63) \\
 & RISLL & 26.7 (1.18) & \textbf{26.5} (1.17) & 26.6 (1.28) & 26.9 (1.20) \\
$n=1500$ & RCLL & 1.58 (0.06) & \textbf{1.57} (0.06) & 1.59 (0.06) & 1.60 (0.07) \\
$p=4$ & ISBS & \textbf{8.5} (0.38) & \textbf{8.5} (0.35) & 8.6 (0.34) & 8.6 (0.58)\\
\bottomrule
\end{tabular}
}
\end{table}

\textbf{Results}\, In \cref{tab:resultsBM3}, we observe that for both, training and evaluation, using different scoring rules only leads to minor changes.
Choosing ISBS as the evaluation metric seems to give a slight advantage to the model which is also trained on an ISBS loss. 
Among the proper scoring rules, we do not observe a similar pattern.


\section{DISCUSSION \& CONCLUSION}

We proposed a new method for estimating event time distributions from censored data, including competing risks using scoring rules as loss function. 
Our framework can be seamlessly integrated into neural networks but also in tree-based models, generalized additive models, and gradient boosting for special cases.
Empirical results demonstrate that the proposed integration of scoring rules yields good predictive performance and the proposed framework is often on par with or outperforms other approaches.

\paragraph{Limitations and Future Work} 
While our approach works for right-censored data and also in the case of competing risks, other SA use cases such as interval-censored data, multi-state modeling, or recurrent events are contemporary challenges that could be an interesting extension of our proposal for future research.
Our method allows the use of proper and improper scoring rules.
Our approach is designed to be agnostic to this.
However, it can be highly beneficial for practitioners to study the choice of the scoring rule more rigorously than we did.
Particularly, our analysis is short of a systematic study on when which scoring rule may be theoretically and practically appropriate.


\bibliographystyle{plainnat}
\bibliography{bibliography}

\begin{thebibliography}{28}
\providecommand{\natexlab}[1]{#1}
\providecommand{\url}[1]{\texttt{#1}}
\expandafter\ifx\csname urlstyle\endcsname\relax
  \providecommand{\doi}[1]{doi: #1}\else
  \providecommand{\doi}{doi: \begingroup \urlstyle{rm}\Url}\fi

\bibitem[Aalen(1978)]{aalen.nonparametric.1978}
Odd Aalen.
\newblock Nonparametric {Inference} for a {Family} of {Counting} {Processes}.
\newblock \emph{The Annals of Statistics}, 6\penalty0 (4), July 1978.

\bibitem[Archetti et~al.(2024)Archetti, Stranieri, and Matteucci]{archettiBridgingGapImprove2024}
Alberto Archetti, Francesco Stranieri, and Matteo Matteucci.
\newblock Bridging the gap: Improve neural survival models with interpolation techniques.
\newblock \emph{Progress in Artificial Intelligence}, September 2024.
\newblock ISSN 2192-6360.
\newblock \doi{10.1007/s13748-024-00343-y}.

\bibitem[Avati et~al.(2020)Avati, Duan, Zhou, Jung, Shah, and Ng]{avati2018}
Anand Avati, Tony Duan, Sharon Zhou, Kenneth Jung, Nigam~H Shah, and Andrew~Y Ng.
\newblock Countdown regression: sharp and calibrated survival predictions.
\newblock In \emph{Uncertainty in Artificial Intelligence}, pages 145--155. PMLR, 2020.

\bibitem[Bender and Scheipl(2018)]{pammtools}
Andreas Bender and Fabian Scheipl.
\newblock pammtools: {Piece}-wise exponential {Additive} {Mixed} {Modeling} tools.
\newblock \emph{arXiv:1806.01042 [stat]}, June 2018.
\newblock arXiv: 1806.01042.

\bibitem[Cox(1972)]{cox.regression.1972}
D.~R. Cox.
\newblock Regression {Models} and {Life}-{Tables}.
\newblock \emph{Journal of the Royal Statistical Society. Series B (Methodological)}, 34\penalty0 (2):\penalty0 187--220, 1972.

\bibitem[Curtis et~al.(2012)Curtis, Shah, Chin, Turashvili, Rueda, Dunning, Speed, Lynch, Samarajiwa, Yuan, et~al.]{curtis2012genomic}
Christina Curtis, Sohrab~P Shah, Suet-Feung Chin, Gulisa Turashvili, Oscar~M Rueda, Mark~J Dunning, Doug Speed, Andy~G Lynch, Shamith Samarajiwa, Yinyin Yuan, et~al.
\newblock The genomic and transcriptomic architecture of 2,000 breast tumours reveals novel subgroups.
\newblock \emph{Nature}, 486\penalty0 (7403):\penalty0 346--352, 2012.

\bibitem[Dawid and Musio(2014)]{dawid2014}
Alexander~Philip Dawid and Monica Musio.
\newblock Theory and applications of proper scoring rules.
\newblock \emph{Metron}, 72\penalty0 (2):\penalty0 169--183, 2014.

\bibitem[Friedman(1982)]{Friedman1982}
Michael Friedman.
\newblock Piecewise exponential models for survival data with covariates.
\newblock \emph{The Annals of Statistics}, 10\penalty0 (1):\penalty0 101--113, 1982.

\bibitem[Gensheimer and Narasimhan(2019)]{gensheimer.scalable.2019}
Michael~F. Gensheimer and Balasubramanian Narasimhan.
\newblock A scalable discrete-time survival model for neural networks.
\newblock \emph{PeerJ}, 7:\penalty0 e6257, January 2019.

\bibitem[Gneiting and Raftery(2007)]{gneiting2007strictly}
Tilmann Gneiting and Adrian~E Raftery.
\newblock Strictly proper scoring rules, prediction, and estimation.
\newblock \emph{Journal of the American statistical Association}, 102\penalty0 (477):\penalty0 359--378, 2007.

\bibitem[Graf et~al.(1999)Graf, Schmoor, Sauerbrei, and Schumacher]{graf1999assessment}
Erika Graf, Claudia Schmoor, Willi Sauerbrei, and Martin Schumacher.
\newblock Assessment and comparison of prognostic classification schemes for survival data.
\newblock \emph{Statistics in medicine}, 18:\penalty0 2529--2545, 1999.

\bibitem[Harrell et~al.(1982)Harrell, Califf, Pryor, Lee, and Rosati]{harrell1982evaluating}
Frank~E Harrell, Robert~M Califf, David~B Pryor, Kerry~L Lee, and Robert~A Rosati.
\newblock Evaluating the yield of medical tests.
\newblock \emph{Jama}, 247\penalty0 (18):\penalty0 2543--2546, 1982.

\bibitem[Hartl et~al.(2022)Hartl, Kopper, Bender, Scheipl, Day, Elke, and K{\"u}chenhoff]{hartl2022protein}
Wolfgang~H Hartl, Philipp Kopper, Andreas Bender, Fabian Scheipl, Andrew~G Day, Gunnar Elke, and Helmut K{\"u}chenhoff.
\newblock Protein intake and outcome of critically ill patients: analysis of a large international database using piece-wise exponential additive mixed models.
\newblock \emph{Critical Care}, 26\penalty0 (1):\penalty0 1--12, 2022.

\bibitem[Howard et~al.(2017)Howard, Chiu, McDonald, Kan, and Yianchen]{kkbox}
Addison Howard, Arden Chiu, Mark McDonald, Wendy Kan, and Yianchen.
\newblock Wsdm - kkbox's music recommendation challenge, 2017.
\newblock URL \url{https://kaggle.com/competitions/kkbox-music-recommendation-challenge}.

\bibitem[Jaeger et~al.(2019)Jaeger, Long, Long, Sims, Szychowski, Min, Mcclure, Howard, and Simon]{jaeger.oblique.2019}
Byron~C. Jaeger, D.~Leann Long, Dustin~M. Long, Mario Sims, Jeff~M. Szychowski, Yuan-I. Min, Leslie~A. Mcclure, George Howard, and Noah Simon.
\newblock Oblique random survival forests.
\newblock \emph{The Annals of Applied Statistics}, 13\penalty0 (3):\penalty0 1847--1883, 2019.

\bibitem[Katzman et~al.(2018)Katzman, Shaham, Cloninger, Bates, Jiang, and Kluger]{katzman2018deepsurv}
Jared~L Katzman, Uri Shaham, Alexander Cloninger, Jonathan Bates, Tingting Jiang, and Yuval Kluger.
\newblock Deepsurv: personalized treatment recommender system using a cox proportional hazards deep neural network.
\newblock \emph{BMC medical research methodology}, 18\penalty0 (1):\penalty0 1--12, 2018.

\bibitem[Kyle et~al.(2002)Kyle, Therneau, Rajkumar, Offord, Larson, Plevak, and Melton]{kyle.long-term.2002}
Robert~A. Kyle, Terry~M. Therneau, S.~Vincent Rajkumar, Janice~R. Offord, Dirk~R. Larson, Matthew~F. Plevak, and L.~Joseph Melton.
\newblock A {Long}-{Term} {Study} of {Prognosis} in {Monoclonal} {Gammopathy} of {Undetermined} {Significance}.
\newblock \emph{New England Journal of Medicine}, 346\penalty0 (8):\penalty0 564--569, February 2002.

\bibitem[Lee et~al.(2018)Lee, Zame, Yoon, and {van der Schaar}]{lee.deephit.2018}
Changhee Lee, William~R. Zame, Jinsung Yoon, and Mihaela {van der Schaar}.
\newblock {DeepHit}: {A} {Deep} {Learning} {Approach} to {Survival} {Analysis} {With} {Competing} {Risks}.
\newblock In \emph{23nd {AAAI} {Conference} on {Artificial} {Intelligence}}, 2018.

\bibitem[Rindt et~al.(2022)Rindt, Hu, Steinsaltz, and Sejdinovic]{rindt2022}
David Rindt, Robert Hu, David Steinsaltz, and Dino Sejdinovic.
\newblock Survival regression with proper scoring rules and monotonic neural networks.
\newblock In \emph{International Conference on Artificial Intelligence and Statistics}, pages 1190--1205. PMLR, 2022.

\bibitem[Schumacher et~al.(1994)Schumacher, Bastert, Bojar, H{\"u}bner, Olschewski, Sauerbrei, Schmoor, Beyerle, Neumann, and Rauschecker]{schumacher1994randomized}
M~Schumacher, G~Bastert, H~Bojar, K~H{\"u}bner, M~Olschewski, W~Sauerbrei, C~Schmoor, C~Beyerle, RL~Neumann, and HF~Rauschecker.
\newblock Randomized 2 x 2 trial evaluating hormonal treatment and the duration of chemotherapy in node-positive breast cancer patients. german breast cancer study group.
\newblock \emph{Journal of Clinical Oncology}, 12\penalty0 (10):\penalty0 2086--2093, 1994.

\bibitem[Sonabend(2021)]{sonabend2021}
Raphael Sonabend.
\newblock \emph{A theoretical and methodological framework for machine learning in survival analysis: Enabling transparent and accessible predictive modelling on right-censored time-to-event data}.
\newblock PhD thesis, UCL (University College London), 2021.

\bibitem[Sonabend et~al.(2022)Sonabend, Zobolas, Kopper, Burk, and Bender]{sonabend2022proper}
Raphael Sonabend, John Zobolas, Philipp Kopper, Lukas Burk, and Andreas Bender.
\newblock Examining properness in the external validation of survival models with squared and logarithmic losses.
\newblock \emph{arXiv preprint arXiv:2212.05260}, 2022.

\bibitem[Tern{\`e}s et~al.(2017)Tern{\`e}s, Rotolo, Heinze, and Michiels]{ternes.identification.2017}
Nils Tern{\`e}s, Federico Rotolo, Georg Heinze, and Stefan Michiels.
\newblock Identification of biomarker-by-treatment interactions in randomized clinical trials with survival outcomes and high-dimensional spaces.
\newblock \emph{Biometrical Journal}, 59\penalty0 (4):\penalty0 685--701, 2017.
\newblock ISSN 1521-4036.
\newblock \doi{10.1002/bimj.201500234}.

\bibitem[Tutz and Schmid(2016)]{tutzModelingDiscreteTimeEvent2016}
Gerhard Tutz and Matthias Schmid.
\newblock \emph{Modeling {{Discrete Time-to-Event Data}}}.
\newblock Springer {{Series}} in {{Statistics}}. Springer International Publishing, Cham, 2016.
\newblock ISBN 978-3-319-28156-8 978-3-319-28158-2.

\bibitem[Wang et~al.(2019)Wang, Li, and Reddy]{wang.machine.2019}
Ping Wang, Yan Li, and Chandan~K. Reddy.
\newblock Machine {Learning} for {Survival} {Analysis}: {A} {Survey}.
\newblock \emph{ACM Computing Surveys (CSUR)}, 51\penalty0 (6):\penalty0 110:1--110:36, February 2019.

\bibitem[Wei(1992)]{wei1992}
Lee-Jen Wei.
\newblock The accelerated failure time model: a useful alternative to the cox regression model in survival analysis.
\newblock \emph{Statistics in medicine}, 11\penalty0 (14-15):\penalty0 1871--1879, 1992.

\bibitem[Wiegrebe et~al.(2024)Wiegrebe, Kopper, Sonabend, Bischl, and Bender]{wiegrebe2024deep}
Simon Wiegrebe, Philipp Kopper, Raphael Sonabend, Bernd Bischl, and Andreas Bender.
\newblock Deep learning for survival analysis: a review.
\newblock \emph{Artificial Intelligence Review}, 57\penalty0 (3):\penalty0 65, 2024.

\bibitem[Yanagisawa(2023)]{yanagisawaProperScoringRules2023}
Hiroki Yanagisawa.
\newblock Proper scoring rules for survival analysis.
\newblock In \emph{International Conference on Machine Learning}, pages 39165--39182. PMLR, 2023.

\end{thebibliography}


\clearpage

\end{document}